%% file: ordinalBucketing.tex
\documentclass[conference]{IEEEtran}
\IEEEoverridecommandlockouts
\usepackage{cite}
\usepackage{amsmath,amssymb,amsfonts}
\usepackage{algorithm}
\usepackage{algorithmic}
\usepackage{graphicx}
\usepackage{textcomp}
\usepackage{xcolor}
\def\BibTeX{{\rm B\kern-.05em{\sc i\kern-.025em b}\kern-.08em
    T\kern-.1667em\lower.7ex\hbox{E}\kern-.125emX}}
\usepackage{tikz}
\usepackage{nicefrac}
\usepackage{pgfplots}
\pgfplotsset{compat=1.13}
\usetikzlibrary{pgfplots.groupplots}

\newcommand{\game}[1]{\textsl{#1}}

\begin{document}

\title{Ordinal Bucketing for Game Trees using \\Dynamic Quantile Approximation
}

\author{\IEEEauthorblockN{
Tobias Joppen}
\IEEEauthorblockA{\textit{Knowledge Engineering Group} \\
\textit{TU Darmstadt}\\
Darmstadt, Germany \\
tjoppen@ke.tu-darmstadt.de}
\and
\IEEEauthorblockN{
Tilman Str\"ubig}
\IEEEauthorblockA{
\textit{TU Darmstadt}\\
Darmstadt, Germany \\
tilman.struebig@googlemail.com}
\and
\IEEEauthorblockN{
Johannes F\"urnkranz}
\IEEEauthorblockA{\textit{Knowledge Engineering Group} \\
\textit{TU Darmstadt}\\
Darmstadt, Germany \\
juffi@ke.tu-darmstadt.de}
}
\IEEEpubid{\begin{minipage}{\textwidth}\ \\[12pt]
978-1-7281-1884-0/19/\$31.00 \copyright 2019 IEEE
\end{minipage}}
\maketitle

\begin{abstract}
In this paper, we present a 
simple and cheap ordinal bucketing algorithm that approximately generates $q$-quantiles 
from an incremental data stream.
The bucketing is done dynamically in the sense that the amount of buckets $q$ increases with the number of seen samples.
We show how this can be used in Ordinal Monte Carlo Tree Search (OMCTS) to yield better bounds on time and space complexity, especially in the presence of noisy rewards.
%
Besides complexity analysis and quality tests of quantiles, we evaluate our method using OMCTS 
in the General Video Game Framework (GVGAI).
Our results demonstrate
its dominance over vanilla Monte Carlo Tree Search in the presence of noise, where OMCTS without bucketing has a very bad time and space complexity. 

\end{abstract}


\section{Introduction}

Ordinal data are widely used in many real-world scenarios such as in ratings or questionnaires.
In many cases, the set of possible values is limited to a low number of ordinal values, such as 1 to 5 stars, but this is not necessary the case.
The basic assumption of ordinal data is that nothing but the ordering of the values is known.
In particular, no distance measure can be assumed.
This implies, e.g., that averaging, adding or multiplying data values is impossible, in contrast to common real-valued data.
Therefore, ordinal data are much more difficult
to handle than real-valued data, 
which is one reason why they are often interpreted as numerical values.

For example, the framework for the General Video Game AI (GVGAI) competitions, which we use in this paper for evaluation purposes,
includes a large variety of games that can be played.
For doing so,
it provides game playing agents with a numerical score for the given state of a game.
Increasing the score often correlates to performing well and approaching the goal.
Examples for actions that lead to an increase in score include collecting diamonds, catching or slaying enemies, solving a minor puzzle or detecting a key for a door.
With few exceptions and across all games, such events increase the score by exactly one point.
It is likely, that this is not a meaningful distance measure but only an indication of success, or a number to derive an ordering or preferences over states.
Hence, there are arguments to view those scores as ordinal values.

Monte Carlo Tree Search, a basic algorithm often used in GVGAI agents, interprets these scores as real-valued feedback \cite{Yolobot, MCTS}.
In previous work, we have proposed Ordinal-MCTS (OMCTS), an MCTS variant that treats these scores in an ordinal fashion \cite{OMCTS}.
The OMCTS algorithm has a linear factor to time and space complexity dependent on the number of ordinal rewards seen.
This is sufficient for domains with a low number of possible ordinal values, but may become excessive in comparison to MCTS once this number rises.

In this paper, we present an algorithm that uses
bucketing for bounding the number of ordinal values to make OMCTS
work efficient with any stream of ordinal values.
This is a problem especially in settings with noisy reward signals.
We investigate this setting by applying artificial noise to GVGAI games.
Due to the fact that OMCTS spans a tree of game states, and uses bucketing in each of those states, it should be fast and performant for 
any amount of data with as little overhead as possible.

One property of MCTS is the \emph{asymmetric growth} of its search tree.
Actions that lead to better states are visited and explored more often.
Hence, MCTS spends more time searching for good solutions and less time in less interesting parts of a game tree.
Similarly,
one does not want to spend a lot of time or overhead to bucket the ordinal rewards in bad states.
Instead, one is fine with having a coarser approximation of the seen rewards for non-optimal states.
This is in contrast to the well explored parts of the search tree:
Here, one wants to have a fine-grained bucketing to be able to identify the very best action, and thus one is willing to put more time for creating this bucketing.
Since we do not know a priori whether a given action needs a fine-grained bucketing or a rough approximation, we are in need for a dynamic method that improves its quality the more data is seen.
For example,
compare the root node with a newly expanded node:
The root node is frequently visited and needs very detailed information about its reward distribution whereas a node that has just been generated does not need any bucketing at all.

In the following chapter, we start with a discussion of related work that also focuses on reducing the size of a set by merging values or identifying meta concepts, and relate ordinal bucketing to quantile approximation.

\section{Reducing the Cardinality}
The idea of summarizing many ordinal values to a fixed number of bins or buckets
is related to many different aspects of machine learning and statistics, which we briefly survey in the following.

We start with data bucketing or binning: the concept of merging multiple value-quantity pairs to fewer interval-quantity pairs.
In statistics, creating optimal histograms is a well explored research area.
Concepts of optimality are well defined and optimal solutions are known for different error measures \cite{Jagadish}. 
They can often can be obtained in a \emph{first-collect-then-bucket} fashion, where one has to create a histogram for a given distribution sample, or in a streamed way where the bucketing is incrementally updated \cite{Jagadish, BenHaim}.
In this paper, we take a look at a featureless and ordinal way of the latter case.
This is an important point, since common bucketing solutions require and exploit features or a metric over the objects to bin, which makes it 
easy to determine which values are close to each other so that their bins can be merged.
An ordinal scale, however, does not have such a metric.
One cannot tell whether two ordinals are far away or close by, only which of them has a higher value.
There also are ordinal clustering methods that do not require a metric but make use of features\cite{Ranalli}.

Defining the quality of a bucketing
is not trivial without having a metric.
A reasonable idea is to strive for buckets of equal size, which 
leads to $q$-quantiles that split a distribution into $q$ equally sized parts, where each quantile contains $\nicefrac{1}{q}$ of the complete distribution.
Looking once more at the root and leaf node example, the root node might want to have its action rewards organized in many quantiles whereas the leaf node is fine with only storing the current median, the 2-quantile.

We propose a simple algorithm to achieve that, where the focus is on very little overhead and a short run-time.
In comparison to other bucketing or quantile approximating algorithms, our approach only stores very little information and often needs to resort to random decisions, which nevertheless leads to good results.
%
We will test our bucketing algorithm in two distinct ways.
First, we test the quantile approximation as a stand-alone algorithm for streamed data and analyze the error on the quantiles, as well as its run time and space complexity for different kinds of distribution functions.
Second, we use the GVGAI Framework to analyze the influence of OMCTS using this approach.
As a baseline we also compare to vanilla MCTS.


\section{Searching with Ordinal Rewards}

\subsection{Ordinal Markov Decision Processes}
Markov Decision Processes (MDPs)
\cite{MDP}
are problems in which an agent has to repeatedly choose one action $a\in A$ from a given set of possibilities.
Once an action is chosen, the agent moves to another state $s\in S$ while receiving a short-term reward $r \in R$.
This other state may be a terminal state (like the end of a game) or has a new set of actions $A(s) \subseteq A$ to choose from..
The agent's task is to find a good policy to maximize some quality measure.
The most prominent measure of success is the \emph{cumulative regret}, i.e., the difference between the sampled and the optimal (expected) reward.

Ordinal MDPs \cite{OMDP} are a variant of this setting, in which the agent observes an ordinal reward signal $o \in Q$ instead of a real-valued reward.
To make sense of ordinal rewards, an qualitative scale $E = \{o_1 \succ o_2, ... \succ o_n\}$ over all $o_i \in Q$ is given.
Since no metric is applicable in $Q$, and therefore rewards can not be trivially aggregated, one has to use other quality measures for OMDPs.

\subsection{Ordinal Monte Carlo Tree Search}
Monte Carlo tree search (MCTS) is a popular algorithm to approximately solve MDPs in real time \cite{MCTS}.
The algorithm iteratively builds up a 
growing model of the game tree. 
One iteration exists of four phases:
The \emph{selection step} starts at the root node of the tree and iteratively chooses one action given historical information about those actions. 
This is often done using the \emph{UCT} formula \cite{UCT}, which trades off actions that are perceived as good (exploitation) and actions that have not been visited often (exploration).
Once the selection reaches a leaf node, the \emph{expansion step} adds one or more child nodes to the tree.
From there on, a \emph{simulation} is started which performs random actions until a terminal node is found or a computational limit is reached.
In the last step, the (heuristic) value of this final state is used to update the information of all actions along the chosen path, which may change the selected actions in the next iteration.

In previous work, we have introduced an variant of this algorithm for OMDPS, in which the values obtained at the end of each iteration are on an ordinal scale \cite{OMCTS}. 
Its key ingredient is a
relative dominance measure to rate actions for OMDPs.
Here, an action is rated in comparison to its alternatives:
\begin{equation}
B(a) = \frac{1}{|A|-1}\sum_{b;b\neq a}^A \Pr[a \succ b],
\label{formula:B}
\end{equation}
where $A$ is the set of possible actions, $a\in A$ is the action to inspect and $\Pr[a \succ b]$ is the tie-normalized chance of $a$ beating $b$ given two random samples from those arms.
This probability can be estimated empirically.
Current methods have time and space complexities that are linear in the size of $Q$, the set of reward signals \cite{OMCTS}.
Once this set grows out of decent bounds or maybe even becomes infinite (e.g. in the presence of noise) the estimation of $B$ becomes too costly. 
The method proposed in this paper can bound the complexity by a fixed or logarithmic growing maximum number $q$ of buckets, so that an obtained reward does not have to be compared to all
previously observed values, but only to 
$q-1$ points given by the observed $q$-quantiles.

OMCTS uses \eqref{formula:B} as the exploitation term in a modified UCT formula:
\begin{equation}
a^* = \arg\max_{a \in A} B(a) + 2 C \sqrt{\frac{2 \ln n}{n_a}},
\label{eq:o-mcts}
\end{equation}
where $n$ is the number of actions played, $n_a$ is the number of action $a$ played so far and $a^*$ is the next action to choose in the selection step.

The next section introduces three ordinal bucketing methods.
In Section~\ref{sec:integration}, we will show how these can be integrated into OMCTS.

\section{Ordinal Bucketing}
\label{sec:bucketing}
In the following section, we describe three bucketing algorithms with different characteristics and how to derive quantile approximations.
We first introduce the formal problem and the used bucketing structure.

\subsection{Problem Definition}
Given an unknown distribution of objects in an ordinal domain $Q$ represented by a random variable $X$  and a time step $t$, the task is to create a set $H^X_t$ of buckets that bracket all past samples $\hat{X_t} = (X_0, X_1, ... X_t)$ together.
The number of buckets $|H^X_t|\leq f(t)$ has an upper limit defined by a bound function $f(t)\in \mathbb{N}$ which is naturally smaller than $t$ to have a need for a real bucketing.
At any given time $t$, the task is to create a bucketing $H_t^X$ given the previous bucketing $H^X_{t-1}$ and the observation $X_t$.
The algorithm proceeds in an on-line manner, i.e., it is not possible to access all past samples $\hat{X_h}$, but only the previous bucketing which has to be updated.

A \emph{bucket} $g=(g_u,g_n,g_d)$ is defined by an upper bound $g_u \in Q$, a number $g_n \in \mathbb{N}$ and auxiliary data $g_d$ that can be used to calculate a pivot point of this bucket using a globally defined function $P(g_d) \in Q$.
The semantic is that approximately $g_n$ elements of $\hat{X_t}$ lie between $g_u$ and the upper bound of the bucket below $g'_u$, where approximately $\nicefrac{g_n}{2}$ of the buckets in $g$ are above and below the pivot $P(g_d)$ respectively.
Recall that one can not simply interpolate between the min and max value since we are on an ordinal scale.
%

The main idea of our dynamically adapting method is that once a bucket reaches an upper limit on $g_n$, it is split into two buckets, using $P(g_d)$ as their border, or, if the number of possible buckets increases, the largest bucket is split in half.
As an example for bucketing, if one wants to calculate the $2$-quantile (median) of a data stream, you use two buckets. 
The upper bound of the lower bucket represents the median.
Asking for a $3$-quantile, one needs three buckets, and so on.

\subsection{Bucketing Algorithms}
In the following, we explain three novel ordinal bucketing methods.
Every value $o\in Q$ is assigned to exactly one bucket $g_o$.
For convenience we introduce the following
notations:
$N(o)$ is the number of stored values of $g_o$, $U(o)$ is its upper bound and $D(o)$ are the data stored in $g_o$.
%
%
%
The number of stored elements $N(o)$ of $g_o$ is updated by calling 
'\emph{Store $o$}'.

\subsubsection{First-n-Bucketing}
A simple algorithm for bucketing ordinal values takes the first $n$ distinct ordinal values and uses them as upper bounds for its buckets.
Hence, the upper bound function is independent of the number of seen samples: $f(t)=n$.
We call this approach \textbf{First-n-Bucketing} (see Algorithm~\ref{alg:first-n-bucketing}).
It is not capable of dynamically increasing the number of buckets and will be used as a baseline later. 
This algorithm does not store any auxiliary data $g_d$.

\begin{algorithm}[ht]
\caption{Adding a value with \textbf{First-n-Bucketing}}
\label{alg:first-n-bucketing}
\begin{algorithmic}
\REQUIRE Time $t$, Sample $X_t$, Previous Bucketing $H^X_{t-1}$, Number of Buckets $n$
\IF{$|H^X_{t-1}| < n$ \AND $U(X_t) \neq X_t$} 
\STATE $H^X_t = H^X_{t-1} \cup \{(X_t,0,\emptyset)\}$
\ENDIF
\STATE Store $X_t$
\end{algorithmic}
\end{algorithm}

\subsubsection{k-Log-Growing}
The next idea addresses the dynamically increasing number of buckets.
Here, the number of buckets have a logarithmic bound on the number of seen samples: $f(t) = k \log(t) $ with $k$ being a parameter to scale the number of buckets.
We named the resulting algorithm \textbf{k-Log-Growing} (see Algorithm~\ref{alg:k-log-growing-bucketing}).

\begin{algorithm}[b]
\caption{Adding an ordinal value with k-Log-Growing}
\label{alg:k-log-growing-bucketing}
\begin{algorithmic}
\REQUIRE Time $t$, Sample $X_t$, Previous Bucketing $H^X_{t-1}$, Parameter $k$
\IF{$t == 0$}
\STATE Initialize with an empty bucket
\ENDIF
\STATE Store $X_t$
\IF{$|H^X_{t-1}| < k \log (t)$ \AND largest bucket has pivot}
\STATE Split largest bucket
\ENDIF
\end{algorithmic}
\end{algorithm}

In the initialization phase, an empty bucket spanning the complete range is added.
At the beginning, new samples are added to this one bucket.
Once the upper bound of available buckets increases, a new bucket can be added.
Instead of adding a new empty bucket, we split an existing bucket using a pivot point.

For computing the pivot point of 
a bucket $g=(g_u, g_n, g_d)$, this algorithm uses the auxiliary data $g_d\in Q^m$ to store the last $m$ values seen in this bucket, where $m$ is odd.
These $m$ data points can be used to compute an approximate median.
Empirically, we have found 
that $m=3$ is enough to show a decent behavior.
If a bucket has not yet seen $m$ data points, the pivot can not be computed.
We refer to this with '\emph{has pivot}' in the following algorithms.
The auxiliary data of a bucket is updated, whenever a new sample is added into the bucket with \emph{Store $X_i$} and replaces the oldest entry.

An obvious choice for the bucket to be split is the largest bucket.
Its pivot point is used as the splitting point, which results in two equally sized buckets, the lower one having the previous pivot point as its upper bound, and the other re-using the previous upper bound.
If the largest bucket has too few seen samples to estimate the pivot, it is not split but one waits until it has enough data to do so.
Since after initialization all elements are added to the first bucket and splitting is only affected by the last $m$ seen samples, it is easily possible to create non optimal splits.
Especially for streams with a low number of different values (which are repeated often) this initialization could result in an arbitrarily bad bucketing.


\subsubsection{k-log-Growing-First-n}
Combining the two previous ideas, we get an algorithm that applies bucketing only after $n$ distinct ordinal values have been observed and then increases the number of buckets, dependent on the number of observed values. This algorithm, \textbf{k-Log-Growing-First-n} (see Algorithm~\ref{alg:k-log-growing-first-n-bucketing}), boosts the accuracy for few observed values, while it is still able to handle large amounts of data.

\begin{algorithm}[ht]
\caption{Adding an ordinal value with k-Log-Growing-First-n}
\label{alg:k-log-growing-first-n-bucketing}
\begin{algorithmic}
\REQUIRE Time $t$, Sample $X_t$, Previous Bucketing $H^X_{t-1}$, Parameter $k$, Parameter $k$
\IF{$|H^X_{t-1}| < n$ \AND $U(X_t) \neq X_t$} 
\STATE $H^X_t = H^X_{t-1} \cup \{(X_t,0,\emptyset)\}$
\ENDIF
\STATE Store $X_t$
\IF{$|H^X_{t-1}| \geq n$ \AND $|H^X_{t-1}| > k  \log (t)$ \AND largest bucket has pivot}
\STATE Split largest bucket
\ENDIF
\end{algorithmic}
\end{algorithm}

\begin{figure}[b]
    \centering
    \begin{tikzpicture}
        \begin{semilogxaxis}[
                xlabel={Number of added values},
                ylabel={Number of buckets},
                tick align=outside,
                xtick={1, 10, 100, 1000, 10000, 100000},
                log basis x=10,
        ]
        
        \addplot [domain=1:50, samples=20, color=blue,]{x};
        \addplot [domain=1:10, samples=20, color=black,]{x};
        \addplot [domain=1:10, samples=20, color=red,]{x};
        \addplot [domain=10:100000, samples=20, color=black,]{10};
        \addplot [domain=10:100, samples=20, color=red,]{10};
        \addplot [domain=100:100000, samples=20, color=red,]{5 * log10(x)};
        
        \addlegendentry{\textbf{No Bucketing}}
        \addlegendentry{\textbf{First-n-Bucketing}}
        \addlegendentry{\textbf{k-Log-Growing-First-n-Bucketing}}
        
        \end{semilogxaxis}
    \end{tikzpicture}
    \caption{Space-complexity for no bucketing, First-n-Bucketing and k-Log-Growing-First-n-Bucketing}
    \label{fig:space-complexity}
\end{figure}
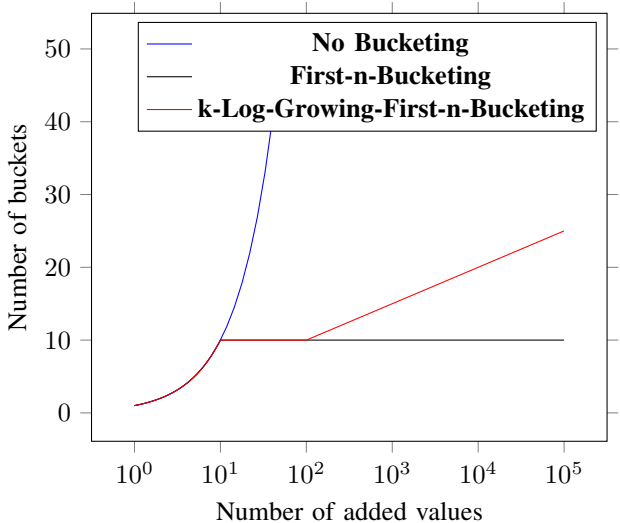

If we take a look at the space-complexity of our presented algorithms (see Fig.~\ref{fig:space-complexity}), the decrease from $\mathcal{O}(t)$ for no bucketing to $\mathcal{O}(\log{}t)$ for \textbf{k-Log-Growing-Bucketing}, respectively $\mathcal{O}(1)$ for \textbf{First-n-Bucketing}, is huge. For time-complexity, we assume a data-structure with logarithmic reading and writing complexity, resulting in $\mathcal{O}(t \log{}t)$ for adding values with no bucketing versus $\mathcal{O}(t \log\log{}t)$ and $\mathcal{O}(t)$ for adding values with \textbf{k-Log-Growing-Bucketing}, respectively \textbf{First-n-Bucketing}. Splitting a single Bucket in \textbf{k-Log-Growing-Bucketing} has a complexity of $\mathcal{O}(\log{}t)$, since it has to iterate over every bucket to find the smallest one, and is performed $\log t$ times, resulting in \textbf{$\mathcal{O}((\log{}t)^2)$}.

%

\section{Analysis of Bucketing Error}

\subsection{Experimental Setup}
We first analyze the performance of our on-line bucketing algorithm. To do so, we 
assume a single stream of ordinal rewards directly sampled from a true distribution.
Our bucketing can not answer queries for a sample probability of a given value $o$, since for each bucket only the stored number of samples, an upper and a lower bound are known.
Hence, it is impossible to measure common error terms like the \emph{Sum of Squared Errors}, a common bucketing error measure.
We instead compare our bucketing to the $q$-quantiles, where $q$ is the number of buckets.
To measure the difference, we look at the $n$-th bucket's upper bound and compare it to the $n$-th $q$-quartile.
Let $u_n$ be the upper bound of bucket $n$, $q_n$ the $n$-th $q$-quartile,  $Rank(o)$ the rank of the value $o$ (the number of samples with a lower value than $o$), and $t$ the complete sample size.
We measure the distance of a bucketing to the $q$-quartiles using:
$$ E(H) = \frac{1}{q} \sum_1^q \frac{|Rank(u_n)-Rank(q_n)|}{t} $$
For the following part of the experiment, we use a Gaussian distribution (unless mentioned otherwise) for sampling and average the results over 100 runs. 
First, we measure $E$ depending on $m$ for different values of $k$ in k-Log-Growing-Bucketing with 1000 sequentially added samples (see Fig.~\ref{fig:m-evaluation}). 
Since $m = 3$ appeared to be a reasonable choice, we use it in the following experiments.
Next, we compare the error for different values of $n$ and $k$ in k-Log-Growing-First-n-Bucketing to see the influence of using the first $n$ observed values as buckets, also averaged over 1000 runs (see Fig.~\ref{fig:n-evaluation}). 
A value of $n = 5$ is used in the upcoming tests. The next experiment examines the behavior of \textbf{k-Log-Growing-First-n-Bucketing} with different values of $k$ for an increasing number of samples (see Fig.~\ref{fig:k-evaluation}), followed by a comparison of our three bucketing algorithms, also dependent on the number of samples. Utilizing the results of the previous experiments, we decided to compare the following configurations: First-n-Bucketing with $n = 5$, k-Log-Growing-Bucketing with $k = 2$ and k-Log-Growing-First-n-Bucketing with $n = 5$ and $k = 2$ (see Fig.~\ref{fig:algorithm-evaluation}). The last experiment uses different distributions to check the performance for different use-cases, like an exponential falling curve, a gaussian and custom defined distribution (see Fig.~\ref{fig:distributions}).
Each distribution is tested for an increasing number of samples to detect potential distribution-dependent behavior of k-Log-Growing-First-n-Bucketing ($n = 5, k = 2$) (see Fig.~\ref{fig:distribution-evaluation}).

\begin{figure}[t!]
    \centering
    \begin{tikzpicture}
        \begin{axis}[
            width=0.95\columnwidth,
            ylabel={Sample Probability},
            xlabel={Object Space},
            ymax=1,
            ymin=0,
            xmin=0,
            xmax=5,
            xticklabels={,,}
            legend style={at={(0.98,0.98)},anchor=north east},
        ]
        
        \addplot [domain=-0:5, samples=20, color=red,]{1/(sqrt(2*pi))*exp(-((x-2.5)^2)/2)};
        
        \addplot [domain=0:5, samples=20, color=blue,]{exp(-x)};
        
         \addplot [domain=0:5, samples=20, color=green,]coordinates{(0, 0)(0, 0.3)(0.25, 0.3)(0.25, 0)(1.25, 0)(1.25, 0.02)(2.5, 0.02)(2.5, 0.5)(2.75, 0.5)(2.75, 0.2)(3, 0.2)(3, 0.125)(5, 0.125) (5, 0)};
        
        \addlegendentry{Gaussian}
        \addlegendentry{Exponential}
        \addlegendentry{Custom}

        \end{axis}
    \end{tikzpicture}
    \caption{The three distributions, used in the experiments.}
    \label{fig:distributions}
\end{figure}
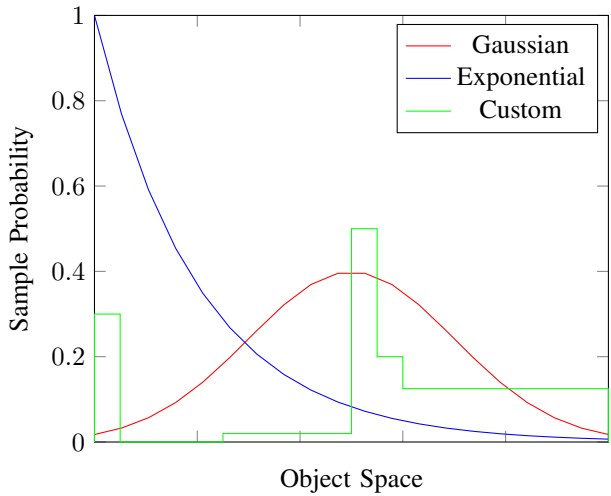

\begin{figure}
    \centering
   
    \begin{tikzpicture}
        \begin{axis}[
                width=0.95\columnwidth,
                xlabel={$m$},
                ylabel={Avg. $E$},
                yticklabel style={
                    /pgf/number format/fixed,
                    /pgf/number format/precision=5
                },
                scaled y ticks=false,
                legend style={at={(0.98,0.98)},anchor=north east}
        ]
        
\addplot [color=red, mark=*,]coordinates{(1, 0.06131857142857144)(3, 0.04531142857142855)(5, 0.04069857142857142)(7, 0.039842857142857145)(9, 0.037910000000000006)(11, 0.03803142857142856)};

\addplot [color=blue, mark=*,]coordinates{(1, 0.04055857142857142)(3, 0.030721428571428584)(5, 0.02931928571428571)(7, 0.027629285714285717)(9, 0.025445000000000002)(11, 0.02369071428571428)};

\addplot [color=yellow, mark=*,]coordinates{(1, 0.03348285714285715)(3, 0.024665714285714287)(5, 0.02262)(7, 0.021217142857142858)(9, 0.020893809523809533)(11, 0.019220000000000008)};

\addplot [color=green, mark=*,]coordinates{(1, 0.027136285714285713)(3, 0.01890971428571428)(5, 0.017731714285714285)(7, 0.017062)(9, 0.015545714285714288)(11, 0.015481428571428573)};

\addplot [color=black, mark=*,]coordinates{(1, 0.021538857142857134)(3, 0.01441057142857143)(5, 0.011829142857142863)(7, 0.010310571428571426)(9, 0.010076)(11, 0.009168878614029449)};
        
        \addlegendentry{$k = 1$}
        \addlegendentry{$k = 2$}
        \addlegendentry{$k = 3$}
        \addlegendentry{$k = 5$}
        \addlegendentry{$k = 10$}
        
        \end{axis}
    \end{tikzpicture}
    \caption{Average distance to the true percentiles for \textbf{k-Log-Growing-Bucketing} with different values of $k$, dependent on $m$. Results for 1000 added values, averaged over 100 runs.}
    \label{fig:m-evaluation}
\end{figure}

\begin{figure}
    \centering
    \begin{tikzpicture}
        \begin{axis}[
                width=0.95\columnwidth,
                xlabel={$n$},
                ylabel={Avg. $E$},
                yticklabel style={
                    /pgf/number format/fixed,
                    /pgf/number format/precision=5
                },
                scaled y ticks=false,
                legend style={at={(0.3,0.98)},anchor=north east}
        ]
        
\addplot [color=red, mark=*,]coordinates{(1, 0.04683428571428571)(2, 0.0510357142857143)(3, 0.06178857142857144)(4, 0.06491428571428572)(5, 0.07876428571428569)(6, 0.08519142857142857)(7, 0.10277714285714282)(8, 0.10799875)(9, 0.09537000000000002)(10, 0.09540900000000001)};

\addplot [color=blue, mark=*,]coordinates{(1, 0.03214285714285714)(2, 0.03140428571428572)(3, 0.033294285714285714)(4, 0.03712142857142858)(5, 0.03530571428571428)(6, 0.03917142857142859)(7, 0.04069071428571429)(8, 0.03928000000000001)(9, 0.042180714285714294)(10, 0.051539285714285725)};

\addplot [color=yellow, mark=*,]coordinates{(1, 0.02474476190476191)(2, 0.024440476190476193)(3, 0.025853809523809522)(4, 0.025768095238095232)(5, 0.02642428571428573)(6, 0.02735380952380952)(7, 0.03009904761904761)(8, 0.02944523809523809)(9, 0.02884761904761903)(10, 0.03253238095238094)};

\addplot [color=green, mark=*,]coordinates{(1, 0.01843885714285715)(2, 0.018932285714285714)(3, 0.019157142857142862)(4, 0.022183142857142856)(5, 0.02086714285714286)(6, 0.020292571428571427)(7, 0.02073085714285714)(8, 0.021096)(9, 0.02358371428571428)(10, 0.02271999999999999)};

\addplot [color=black, mark=*,]coordinates{(1, 0.013607857142857145)(2, 0.015248142857142854)(3, 0.014837285714285718)(4, 0.014951571428571422)(5, 0.015274142857142856)(6, 0.014382857142857142)(7, 0.01483028571428571)(8, 0.01468342857142857)(9, 0.013619428571428567)(10, 0.015491571428571427)};

        \addlegendentry{$k = 1$}
        \addlegendentry{$k = 2$}
        \addlegendentry{$k = 3$}
        \addlegendentry{$k = 5$}
        \addlegendentry{$k = 10$}
        
        \end{axis}
    \end{tikzpicture}
    \caption{Average distance to the true percentiles for \textbf{k-Log-Growing-First-n-Bucketing} with different values of $k$, dependent on $n$. Results for 1000 added values, averaged over 100 runs.}
    \label{fig:n-evaluation}
\end{figure}
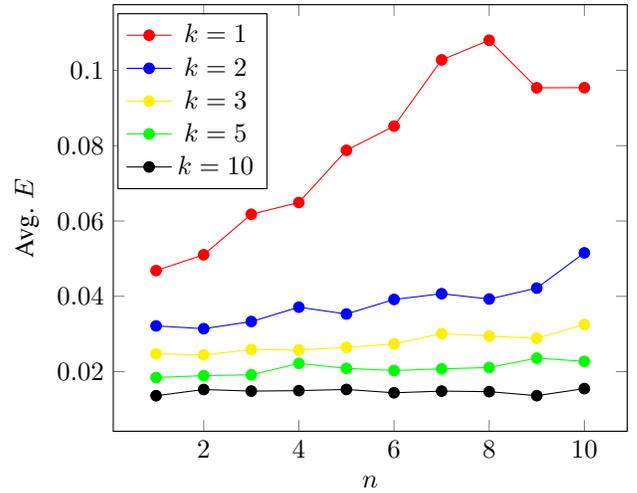

\begin{figure}
    \centering
    
    \begin{tikzpicture}
        \begin{semilogxaxis}[
                width=0.95\columnwidth,
                xlabel={Number of added values},
                ylabel={Avg. $E$},
                yticklabel style={
                    /pgf/number format/fixed,
                    /pgf/number format/precision=5
                },
                scaled y ticks=false,
                legend style={at={(0.98,0.98)},anchor=north east},
        ]
        
\addplot [color=red, mark=*,]coordinates{(10, 0.12239999999999997)(100, 0.12615999999999997)(1000, 0.07444714285714285)(10000, 0.04947790000000001)(100000, 0.039343525000000004)};

\addplot [color=blue, mark=*,]coordinates{(10, 0.11959999999999997)(100, 0.05539000000000001)(1000, 0.037105000000000006)(10000, 0.02793405263157895)(100000, 0.024243441666666678)};

\addplot [color=yellow, mark=*,]coordinates{(10, 0.10023333333333329)(100, 0.04430714285714287)(1000, 0.027314285714285715)(10000, 0.020437821428571434)(100000, 0.018480620000000003)};

\addplot [color=green, mark=*,]coordinates{(10, 0.10339999999999998)(100, 0.027068454733672153)(1000, 0.020391714285714287)(10000, 0.015001468085106379)(100000, 0.014141391379310348)};

\addplot [color=black, mark=*,]coordinates{(10, 0.11226666666666665)(100, 0.02222516732523404)(1000, 0.01544114285714285)(10000, 0.010613602150537632)(100000, 0.010584214655172413)};
        
        \addlegendentry{$k = 1$}
        \addlegendentry{$k = 2$}
        \addlegendentry{$k = 3$}
        \addlegendentry{$k = 5$}
        \addlegendentry{$k = 10$}
        
        \end{semilogxaxis}
    \end{tikzpicture}
    \caption{Average distance to the true percentiles for \textbf{k-Log-Growing-First-n-Bucketing} with different values of $k$, dependent on the number of added values. Results for $n = 5$, averaged over 100 runs.}
    \label{fig:k-evaluation}
\end{figure}
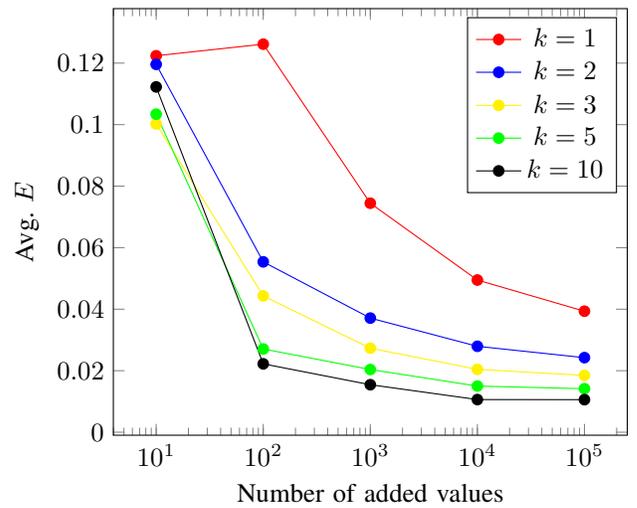

\begin{figure*}
\hfill
\begin{minipage}[t]{0.48\textwidth}
    \centering
    \begin{tikzpicture}
        \begin{semilogxaxis}[
                width=\columnwidth,
                xlabel={Number of added values},
                ylabel={Avg. $E$},
                yticklabel style={
                    /pgf/number format/fixed,
                    /pgf/number format/precision=5
                },
                scaled y ticks=false,
                legend style={at={(0.98,0.65)},anchor=north east},
        ]
        
\addplot [color=red, mark=*,]coordinates{(10, 0.12759999999999996)(100, 0.12286000000000002)(1000, 0.12024199999999997)(10000, 0.137309)(100000, 0.12857317999999998)};

\addplot [color=blue, mark=*,]coordinates{(10, 0.0773333333333333)(100, 0.04174999999999999)(1000, 0.030804285714285746)(10000, 0.02582231578947369)(100000, 0.022637758333333324)};

\addplot [color=green, mark=*,]coordinates{(10, 0.12819999999999993)(100, 0.05016999999999999)(1000, 0.03359857142857143)(10000, 0.03054031578947369)(100000, 0.023216312500000003)};
        
        \addlegendentry{\textbf{First-5}}
        \addlegendentry{\textbf{2-Log-Growing}}
        \addlegendentry{\textbf{2-Log-Growing-First-5}}
        
        \end{semilogxaxis}
    \end{tikzpicture}
    \caption{Average distance to the true percentiles for instances of the three presented algorithms, dependent on the number of added values. Results averaged over 100 runs.}
    \label{fig:algorithm-evaluation}
\end{minipage}
\hfill
\begin{minipage}[t]{0.48\textwidth}
    \centering
    \begin{tikzpicture}
        \begin{semilogxaxis}[
                width=\columnwidth,
                xlabel={Number of added values},
                ylabel={Avg. $E$},
                yticklabel style={
                    /pgf/number format/fixed,
                    /pgf/number format/precision=5
                },
                scaled y ticks=false,
                legend style={at={(0.98,0.98)},anchor=north east},
        ]
        
\addplot [color=red, mark=*,]coordinates{(10, 0.07841666666666662)(100, 0.041259999999999984)(1000, 0.030067142857142858)(10000, 0.02490305263157895)(100000, 0.024205512500000005)};

\addplot [color=blue, mark=*,]coordinates{(10, 0.07274999999999995)(100, 0.04320999999999999)(1000, 0.029174999999999996)(10000, 0.023299631578947367)(100000, 0.023556537500000006)};

\addplot [color=green, mark=*,]coordinates{(10, 0.07341666666666663)(100, 0.04085999999999998)(1000, 0.03170785714285715)(10000, 0.02651221052631579)(100000, 0.02514498333333334)};
        
        \addlegendentry{Gaussian}
        \addlegendentry{Exponential}
        \addlegendentry{Custom}
        
        \end{semilogxaxis}
    \end{tikzpicture}
    \caption{Average distance to the true percentiles for three different distribution, dependent on the number of added values. Results for \textbf{k-Log-Growing-First-n-Bucketing} with $k = 2$ and $n = 5$, averaged over 100 runs.}
    \label{fig:distribution-evaluation}
    \end{minipage}
    \hspace*{4mm}
\end{figure*}

\subsection{Results}

In these experiments a single bucketing is used to bucket a bigger stream of data (up to $10^5$).
%
Figure \ref{fig:m-evaluation} shows a decrease of the error $E$ from $m = 1$ to $m = 3$, while the behavior for $m > 3$ does increase again for all settings except for $k=1$.
Hence, storing the last three values is a decent choice for the algorithm compared to the other tested alternatives.
As explained in the previous section, the following tests are done using $m=3$.

Figure \ref{fig:n-evaluation} shows that for $k > 1$ the error $E$ is fairly independent of $n$. The sharp increase for $k = 1$ can be explained by the fact, that after 7 or more initial buckets, no further splits are performed for 1000 added samples, resulting in the same behavior as First-n-Bucketing. 
Overall, it seems to be the case that for $n \rightarrow k \log{}T$, where $T$ is the total number of added values, the error increases, which also explains the slight upward trend for $k = 2$ and $n \ge 9$. 
If any assumptions regarding $T$ are possible, this information could be used to tune $n$ respectively.
But in our case, we decided to go for a value of $n = 5$ to separate k-Log-Growing-First-n-Bucketing from the simple k-Log-Growing-Bucketing, while avoiding a strong, $n$-induced increase of error.

The experiments confirm the intuition, that a larger number of buckets, induced by $k$, results in a smaller error $E$ (cf.~Figures \ref{fig:m-evaluation}, \ref{fig:n-evaluation} and \ref{fig:k-evaluation}). Since this behavior was expected, it also justifies the error-measure itself. Figure \ref{fig:k-evaluation} also shows a fairly stable trend, once the first 100 values have been added. The lack of improvement for $k = 1$ between 10 and 100 derives from the same problem ($n \rightarrow k \log{}T$), we described earlier.

Figure \ref{fig:algorithm-evaluation} shows a steady error for the First-n-Bucketing, while the k-Log-Growing-Bucketing and k-Log-Growing-First-n-Bucketing converge to a similar, lower error. The idea of using the first $n$ values as a foundation for further splits doesn't seem to help, since it induces the initial high error of the First-n-Bucketing. This doesn't affect the performance for large amounts of values, but k-Log-Growing-Bucketing generates a strictly lower error, which also matches the results in Figure \ref{fig:n-evaluation}, where no error-decrease could be seen with an increasing $n$.

Finally, Figure~\ref{fig:distribution-evaluation} exhibits an almost identical performance of k-Log-Growing-First-n-Bucketing on all three used distributions, indicating a good robustness.

\section{Integration of Bucketing into OMCTS}
\label{sec:integration}
After defining how we dynamically approximate quantiles for a stream of data, we now show how to integrate bucketing into OMCTS.
To this end, we take a look at how different actions are rated locally in a given node.

In vanilla OMCTS, the complete estimated probability distribution functions $f_a(o)$ for each action $a$ and value $o$ are stored and updated.
Given two actions $a_i$ and $a_j$ one can estimate the probability 
\begin{equation}\Pr(a_i \succ a_j) = \Pr(a_i > a_i) + \frac{1}{2} \Pr(a_i = a_i).
\end{equation}
Hereby, $\Pr(a_i > a_i)$ can be estimated by computing the integral over the sampled $Q$ values using $f$ of $a_i$ and $a_j$ \cite{OMCTS}.
The linear complexity of OMCTS arises from solving this integral and updating $f$ on all sampled values.
Using our bucketing method, we can reduce this complexity.
Instead of storing and iterating over the complete list of seen values, we now only store and iterate over the stored buckets.
For an action $a$ and its bucketing $H^{X_a}$ with $s_a$ buckets and $s_t$ aggregated values, a bucketed distribution function $\hat{f}_a(o)$ can be derived.
In our experiments we have used the upper bound $g_u$ as a representative value for a given bucket $g$ and the relative proportion of this bucket $g_r=\nicefrac{g_n}{s_t}$ as its sample probability.
Therefore, we interpret the buckets as if only the representative value would have been seen with the cumulative sample probability of all values in this bucket.
Hence, for a bucket $g$ it holds that $\hat{f}_a(g_u)=g_r$ and $\hat{f}_a(o)=0$ for all other values $o$.
Finally, we have used $\hat{f}$ instead of $f$ to estimate $\Pr(a_i \succ a_j)$ for the bucketed OMCTS versions in the following experiments.

\section{Experiments on GVGAI Games}

\subsection{Experimental Setup}

First, we analyze the performance of bucketed OMCTS on GVGAI games in comparison to OMCTS without bucketing and plain MCTS.
We compare the quality of play of these algorithms along the following dimensions: win rate, achieved score, and average number of iterations.
The win rate is the most important measure, since the very first task of an game playing agent is to win games.
The second-order task is to win with a high score.
We also inspect the average number of iterations per turn to analyze the run time complexity of different approaches.
To tackle non-determinism, we average those values over 100 experiments.
As default for GVGAI, an agent has 40 milliseconds to choose an action.
Additionally, we also test agents with 200 ms to see the results for higher sample sizes.
Additionally, we repeat these experiments by disturbing the obtained rewards with artificial Gaussian noise with  standard deviations of $0.1$, $1$ and $10$, which essentially has the effect that no reward is seen more than once, and bucketing becomes crucial for a good performance.

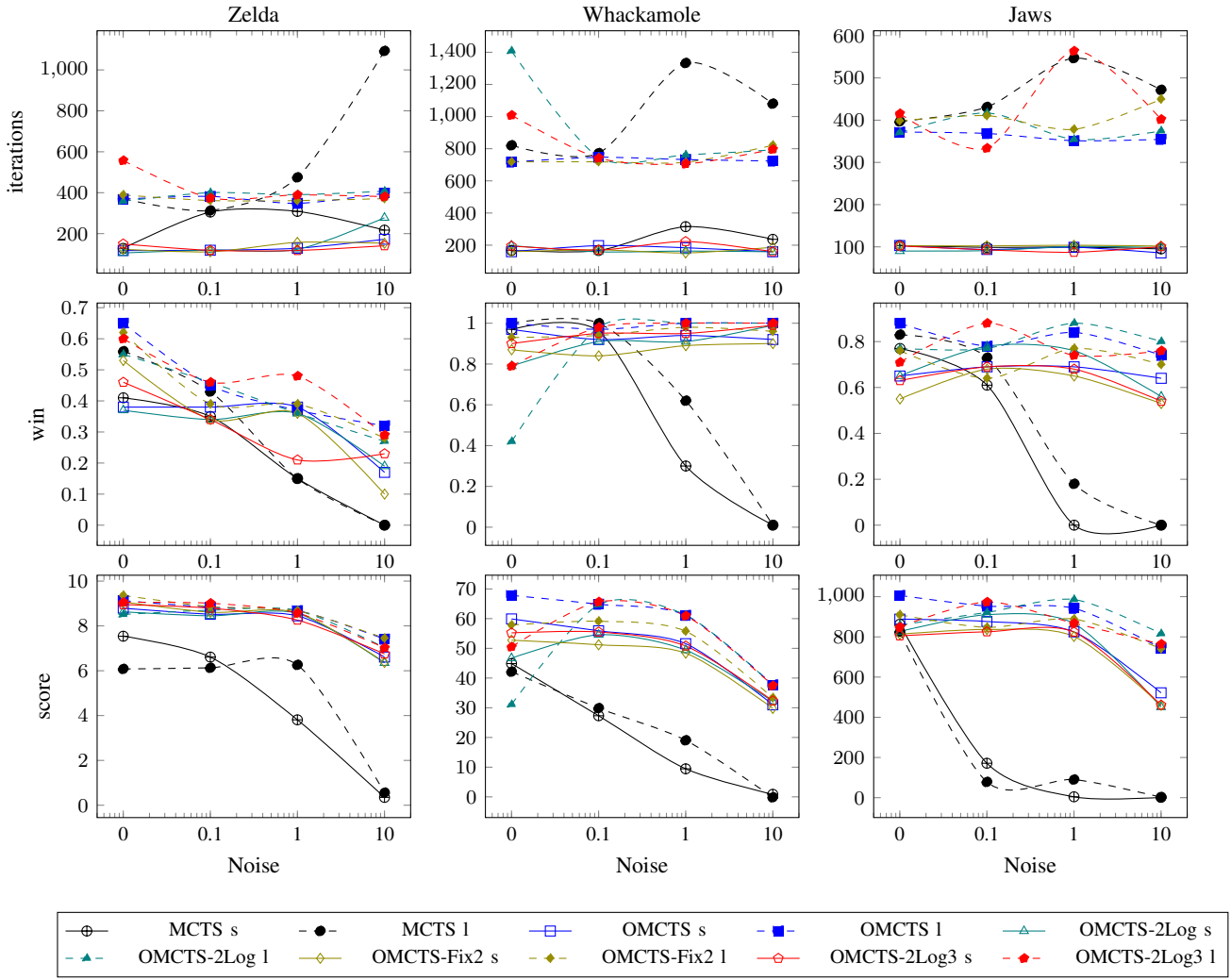
\begin{figure*}[t!]
    \input{winScoreGraph.tex}
    \caption{The results, scores and iterations of all five algorithms on three games having 20ms (\emph{s} postfix) and 400ms (\emph{l} postfix)}
    \label{fig:resultsGVGAI}
\end{figure*}

The tested algorithms are MCTS, OMCTS, and OMCTS with the three different bucketing methods introduced in Section~\ref{sec:bucketing}: OMCTS-Fix2 (using First-2-Bucketing), OMCTS-2Log(using 2-Log-Growing-Bucketing) and OMCTS-2Log3 (using 2-Log-First-3-Growing-Bucketing) as described in the last section.
These five algorithms are tested on three GVGAI games, \game{Zelda}, \game{Whackamole} and \game{Jaws}.
Each of these games has interesting characteristics to test on:
\begin{itemize}
    \item \game{Whackamole} is a quite simple game where one walks around and collects mushrooms that randomly spawn and grant score. 
    There also is a cat one has to avoid since touching it is the only way to lose the game.
    It is easy to avoid this collision.

    \item \game{Jaws} has many enemies that can kill the agent, but done right one can kill them easily and eventually reach a $+1000$ score bonus.

    \item \game{Zelda}, unlike the other games, does not have many score changes while playing the game.
    The player can collect a key ($+1$ points), and can also slay enemies with your sword that kill you on collision ($+2$ points for each killed enemy).
    Interestingly, picking up the key is necessary to win the game whereas killing the enemies is not.
    Having a lot of noise hence might lead to not picking up the key but slaying the enemies, which can result in a lost game with relatively high score.
    
    Finally after $2000$ game ticks, \game{Zelda} is lost if the agent did not manage to collect the key and exit through the door. 
\end{itemize}

\subsection{Results}

Figure~\ref{fig:resultsGVGAI} shows the average number of iterations per turn and the average win and score values per game in dependence of different noise levels.
Over all games, one can see that the number of wins, the score and the number of iterations do not differ significantly for different bucketing versions. Even using no bucketing does not show a significant difference.

The most outstanding result is how the real-valued MCTS and ordinal MCTS versions behave in the presence of noise:
The performance of MCTS constantly drops when the noise is increased having nearly zero wins at a SD of 10.
Even though this is not unexpected at such a high noise level, OMCTS is, on the other hand, still able to perform well. Thus, it seems to be much more robust against noisy rewards because although the obtained score decreases, the winning chances of OMCTS do not suffer as much as those of MCTS. 
In \game{Zelda}, OMCTS seems to struggle to find the key, which, it being a prerequisite for winning the game, results in fewer wins.
For \game{Jaws}, the amount of wins stays unsteady but constantly in the area of $60\%$ to $90\%$.
In \game{Whackamole}, one even can see an increase in wins given noise.
A reason for that is that colliding with the cat can still be evaded (since losing is a high negative reward) and additionally the agent is not lured into dangerous positions (where a mushroom is right next to the cat) since the noise conceals these positive rewards.
The only outstanding point looking at OMCTS variants is the bad performance of OMCTS-2Log in \game{Whackamole}.
As mentioned in Section~\ref{sec:bucketing}, there is a chance of this algorithm to fail due to a bad initialization.
One can see that this does not happen in the presence of noise, since chances of seeing the same reward twice goes to zero.

Looking at the average number of iterations, one can see a direct correlation of iterations and lost games.
Sadly the number of iterations that the agent can perform seem to differ heavily with the current state of the game and whether terminal states are often sampled or not.
Hence, we can not deduce a significant difference of iterations whether bucketing is used or not.
Most of the time the number of iterations and hence also rewards is below $4000$, even for the extended $200$ ms turns.
This number seems to be too small to make a significant time saving in contrast to the expense of the GVGAI framework itself.
Further more, it is even more interesting to see, that for iteration numbers below $500$ the use of bucketing does not decrease the performance.
This could be easily possible since bucketing in each node induces an overhead which only is significant for very low stored samples.

\section{Conclusion}
We have proposed an ordinal bucketing method which separates a stream of data into multiple buckets, where the number of buckets increases
with the amount of available data.
Since ordinal values do not allow the use of distance-based error measures, the bucketing strategy tries to keep the buckets filled equally, leading to quantile estimation.

Our results show that the proposed $k$-log-first-$n$-bucketing has a good runtime and quality of play for both small and large amounts of data.
Using the GVGAI framework we show interesting results comparing OMCTS with and without bucketing:
While both have an overall good performance, OMCTS shows a completely different behavior than MCTS in the presence of noise, where MCTS fails to win games and OMCTS loses score but mainly is able to keep the amount of won games. 

\section*{Acknowledgment}
We gratefully acknowledge the use of the Lichtenberg high performance computer of the TU Darmstadt for our experiments.
This work has been supported by the German Research Foundation (DFG). 

\end{document}

%% file: winScoreGraph.tex
\begin{tikzpicture}
\pgfplotsset{footnotesize,samples=10}
\begin{groupplot}[group style = {group size = 3 by 3, horizontal sep = 30pt, vertical sep = 12pt}, width = 6.0cm, height = 5.0cm]
\nextgroupplot[xmode=log,xticklabels={,0,0.1,1,10}, ylabel=iterations, xlabel=, title = {Zelda}, legend style = { column sep = 10pt, legend columns = 5, legend to name = grouplegend,}]
	\addplot[smooth,black, mark=oplus] plot coordinates { 
		(0.01,127.80253200)
		(0.1,305.37453900)
		(1,308.69443600)
		(10,217.06143100)
	};
	\addlegendentry{MCTS s}

	\addplot[smooth,black, mark=oplus*, dashed] plot coordinates { 
		(0.01,370.62831300)
		(0.1,312.94233700)
		(1,474.82624600)
		(10,1092.47837800)
	};
	\addlegendentry{MCTS l}

	\addplot[smooth,blue, mark=square] plot coordinates { 
		(0.01,117.59467700)
		(0.1,118.80253300)
		(1,128.77517300)
		(10,172.46400300)
	};
	\addlegendentry{OMCTS s}

	\addplot[smooth,blue, mark=square*, dashed] plot coordinates { 
		(0.01,368.69592000)
		(0.1,381.91005900)
		(1,347.94363700)
		(10,397.64826600)
	};
	\addlegendentry{OMCTS l}

	\addplot[smooth,teal, mark=triangle] plot coordinates { 
		(0.01,105.68283200)
		(0.1,119.65311900)
		(1,123.24226900)
		(10,277.65617100)
	};
	\addlegendentry{OMCTS-2Log s}

	\addplot[smooth,teal, mark=triangle*, dashed] plot coordinates { 
		(0.01,358.25876000)
		(0.1,400.14174300)
		(1,390.50637300)
		(10,407.33543600)
	};
	\addlegendentry{OMCTS-2Log l}

	\addplot[smooth,olive, mark=diamond] plot coordinates { 
		(0.01,123.76917200)
		(0.1,111.31124500)
		(1,157.25170300)
		(10,152.89876100)
	};
	\addlegendentry{OMCTS-Fix2 s}

	\addplot[smooth,olive, mark=diamond*, dashed] plot coordinates { 
		(0.01,388.96660000)
		(0.1,361.75490200)
		(1,361.41630600)
		(10,371.08217800)
	};
	\addlegendentry{OMCTS-Fix2 l}

	\addplot[smooth,red, mark=pentagon] plot coordinates { 
		(0.01,149.57756500)
		(0.1,118.29732900)
		(1,117.70764700)
		(10,140.93145500)
	};
	\addlegendentry{OMCTS-2Log3 s}

	\addplot[smooth,red, mark=pentagon*, dashed] plot coordinates { 
		(0.01,557.45701900)
		(0.1,374.16932000)
		(1,389.34861100)
		(10,380.86424100)
	};
	\addlegendentry{OMCTS-2Log3 l}

\nextgroupplot[xmode=log,xticklabels={,0,0.1,1,10}, xlabel=, title = {Whackamole},]
	\addplot[smooth,black, mark=oplus] plot coordinates { 
		(0.01,167.59247800)
		(0.1,166.72955500)
		(1,315.28371300)
		(10,236.45871300)
	};
	\addlegendentry{MCTS short}

	\addplot[smooth,black, mark=oplus*, dashed] plot coordinates { 
		(0.01,821.08488000)
		(0.1,772.55354000)
		(1,1332.97615400)
		(10,1080.11169500)
	};
	\addlegendentry{MCTS long}

	\addplot[smooth,blue, mark=square] plot coordinates { 
		(0.01,159.37374500)
		(0.1,197.16455500)
		(1,183.97793300)
		(10,158.99592800)
	};
	\addlegendentry{OMCTS short}

	\addplot[smooth,blue, mark=square*, dashed] plot coordinates { 
		(0.01,717.25966000)
		(0.1,746.51776500)
		(1,732.44060000)
		(10,723.40614000)
	};
	\addlegendentry{OMCTS long}

	\addplot[smooth,teal, mark=triangle] plot coordinates { 
		(0.01,197.53887000)
		(0.1,158.65706000)
		(1,163.56323400)
		(10,158.83265800)
	};
	\addlegendentry{OMCTS-1 short}

	\addplot[smooth,teal, mark=triangle*, dashed] plot coordinates { 
		(0.01,1407.98080200)
		(0.1,747.34046300)
		(1,761.55274000)
		(10,792.86130000)
	};
	\addlegendentry{OMCTS-1 long}

	\addplot[smooth,olive, mark=diamond] plot coordinates { 
		(0.01,159.08258400)
		(0.1,171.09857300)
		(1,151.73288600)
		(10,186.96055800)
	};
	\addlegendentry{OMCTS-1 short}

	\addplot[smooth,olive, mark=diamond*, dashed] plot coordinates { 
		(0.01,718.16697900)
		(0.1,718.80834800)
		(1,715.42616800)
		(10,821.40922800)
	};
	\addlegendentry{OMCTS-1 long}

	\addplot[smooth,red, mark=pentagon] plot coordinates { 
		(0.01,194.75699100)
		(0.1,169.52802100)
		(1,222.71879600)
		(10,159.44233700)
	};
	\addlegendentry{OMCTS-3 short}

	\addplot[smooth,red, mark=pentagon*, dashed] plot coordinates { 
		(0.01,1008.63271000)
		(0.1,742.07748900)
		(1,705.69798000)
		(10,796.16016000)
	};
	\addlegendentry{OMCTS-3 long}

\legend{};
\nextgroupplot[xmode=log,xticklabels={,0,0.1,1,10}, xlabel=, title = {Jaws},]
	\addplot[smooth,black, mark=oplus] plot coordinates { 
		(0.01,103.25312300)
		(0.1,98.79049600)
		(1,99.56644900)
		(10,95.13371600)
	};
	\addlegendentry{MCTS short}

	\addplot[smooth,black, mark=oplus*, dashed] plot coordinates { 
		(0.01,396.91197300)
		(0.1,431.18546200)
		(1,546.81641900)
		(10,471.75540900)
	};
	\addlegendentry{MCTS long}

	\addplot[smooth,blue, mark=square] plot coordinates { 
		(0.01,103.12973300)
		(0.1,93.84072500)
		(1,99.07474100)
		(10,85.64698200)
	};
	\addlegendentry{OMCTS short}

	\addplot[smooth,blue, mark=square*, dashed] plot coordinates { 
		(0.01,371.78902100)
		(0.1,368.45459700)
		(1,351.43429900)
		(10,354.57411600)
	};
	\addlegendentry{OMCTS long}

	\addplot[smooth,teal, mark=triangle] plot coordinates { 
		(0.01,90.17260100)
		(0.1,91.11956600)
		(1,100.99793600)
		(10,98.27007700)
	};
	\addlegendentry{OMCTS-1 short}

	\addplot[smooth,teal, mark=triangle*, dashed] plot coordinates { 
		(0.01,370.77782200)
		(0.1,417.09750700)
		(1,354.36170400)
		(10,374.54914200)
	};
	\addlegendentry{OMCTS-1 long}

	\addplot[smooth,olive, mark=diamond] plot coordinates { 
		(0.01,102.62099400)
		(0.1,102.26036400)
		(1,103.90658700)
		(10,101.83589000)
	};
	\addlegendentry{OMCTS-1 short}

	\addplot[smooth,olive, mark=diamond*, dashed] plot coordinates { 
		(0.01,398.29547000)
		(0.1,410.60635600)
		(1,378.24487300)
		(10,449.74042300)
	};
	\addlegendentry{OMCTS-1 long}

	\addplot[smooth,red, mark=pentagon] plot coordinates { 
		(0.01,102.25786900)
		(0.1,92.97501100)
		(1,87.21895000)
		(10,100.54391800)
	};
	\addlegendentry{OMCTS-3 short}

	\addplot[smooth,red, mark=pentagon*, dashed] plot coordinates { 
		(0.01,415.48440600)
		(0.1,333.77088200)
		(1,563.78594400)
		(10,401.89336900)
	};
	\addlegendentry{OMCTS-3 long}

\legend{};
\nextgroupplot[xmode=log,xticklabels={,0,0.1,1,10}, ylabel=win, xlabel=, title = {},]
	\addplot[smooth,black, mark=oplus] plot coordinates { 
		(0.01,0.41000)
		(0.1,0.35000)
		(1,0.15000)
		(10,0.00000)
	};
	\addlegendentry{MCTS short}

	\addplot[smooth,black, mark=oplus*, dashed] plot coordinates { 
		(0.01,0.56000)
		(0.1,0.43000)
		(1,0.15000)
		(10,0.00000)
	};
	\addlegendentry{MCTS long}

	\addplot[smooth,blue, mark=square] plot coordinates { 
		(0.01,0.38000)
		(0.1,0.38000)
		(1,0.38000)
		(10,0.17000)
	};
	\addlegendentry{OMCTS short}

	\addplot[smooth,blue, mark=square*, dashed] plot coordinates { 
		(0.01,0.65000)
		(0.1,0.45000)
		(1,0.37000)
		(10,0.32000)
	};
	\addlegendentry{OMCTS long}

	\addplot[smooth,teal, mark=triangle] plot coordinates { 
		(0.01,0.37000)
		(0.1,0.34000)
		(1,0.36000)
		(10,0.19000)
	};
	\addlegendentry{OMCTS-1 short}

	\addplot[smooth,teal, mark=triangle*, dashed] plot coordinates { 
		(0.01,0.55000)
		(0.1,0.46000)
		(1,0.36000)
		(10,0.27000)
	};
	\addlegendentry{OMCTS-1 long}

	\addplot[smooth,olive, mark=diamond] plot coordinates { 
		(0.01,0.53000)
		(0.1,0.34000)
		(1,0.36000)
		(10,0.10000)
	};
	\addlegendentry{OMCTS-1 short}

	\addplot[smooth,olive, mark=diamond*, dashed] plot coordinates { 
		(0.01,0.62000)
		(0.1,0.39000)
		(1,0.39000)
		(10,0.28000)
	};
	\addlegendentry{OMCTS-1 long}

	\addplot[smooth,red, mark=pentagon] plot coordinates { 
		(0.01,0.46000)
		(0.1,0.34000)
		(1,0.21000)
		(10,0.23000)
	};
	\addlegendentry{OMCTS-3 short}

	\addplot[smooth,red, mark=pentagon*, dashed] plot coordinates { 
		(0.01,0.60000)
		(0.1,0.46000)
		(1,0.48000)
		(10,0.29000)
	};
	\addlegendentry{OMCTS-3 long}

\legend{};
\nextgroupplot[xmode=log,xticklabels={,0,0.1,1,10}, xlabel=, title = {},]
	\addplot[smooth,black, mark=oplus] plot coordinates { 
		(0.01,0.97000)
		(0.1,0.96000)
		(1,0.30000)
		(10,0.01000)
	};
	\addlegendentry{MCTS short}

	\addplot[smooth,black, mark=oplus*, dashed] plot coordinates { 
		(0.01,1.00000)
		(0.1,1.00000)
		(1,0.62000)
		(10,0.01000)
	};
	\addlegendentry{MCTS long}

	\addplot[smooth,blue, mark=square] plot coordinates { 
		(0.01,0.97000)
		(0.1,0.92000)
		(1,0.94000)
		(10,0.92000)
	};
	\addlegendentry{OMCTS short}

	\addplot[smooth,blue, mark=square*, dashed] plot coordinates { 
		(0.01,1.00000)
		(0.1,0.97000)
		(1,1.00000)
		(10,1.00000)
	};
	\addlegendentry{OMCTS long}

	\addplot[smooth,teal, mark=triangle] plot coordinates { 
		(0.01,0.79000)
		(0.1,0.91000)
		(1,0.91000)
		(10,0.99000)
	};
	\addlegendentry{OMCTS-1 short}

	\addplot[smooth,teal, mark=triangle*, dashed] plot coordinates { 
		(0.01,0.42000)
		(0.1,0.98000)
		(1,1.00000)
		(10,1.00000)
	};
	\addlegendentry{OMCTS-1 long}

	\addplot[smooth,olive, mark=diamond] plot coordinates { 
		(0.01,0.87000)
		(0.1,0.84000)
		(1,0.89000)
		(10,0.90000)
	};
	\addlegendentry{OMCTS-1 short}

	\addplot[smooth,olive, mark=diamond*, dashed] plot coordinates { 
		(0.01,0.93000)
		(0.1,0.94000)
		(1,0.98000)
		(10,0.96000)
	};
	\addlegendentry{OMCTS-1 long}

	\addplot[smooth,red, mark=pentagon] plot coordinates { 
		(0.01,0.90000)
		(0.1,0.95000)
		(1,0.95000)
		(10,0.99000)
	};
	\addlegendentry{OMCTS-3 short}

	\addplot[smooth,red, mark=pentagon*, dashed] plot coordinates { 
		(0.01,0.79000)
		(0.1,0.98000)
		(1,1.00000)
		(10,1.00000)
	};
	\addlegendentry{OMCTS-3 long}

\legend{};
\nextgroupplot[xmode=log,xticklabels={,0,0.1,1,10}, xlabel=, title = {},]
	\addplot[smooth,black, mark=oplus] plot coordinates { 
		(0.01,0.77000)
		(0.1,0.61000)
		(1,0.00000)
		(10,0.00000)
	};
	\addlegendentry{MCTS short}

	\addplot[smooth,black, mark=oplus*, dashed] plot coordinates { 
		(0.01,0.83000)
		(0.1,0.73000)
		(1,0.18000)
		(10,0.00000)
	};
	\addlegendentry{MCTS long}

	\addplot[smooth,blue, mark=square] plot coordinates { 
		(0.01,0.65000)
		(0.1,0.69000)
		(1,0.69000)
		(10,0.64000)
	};
	\addlegendentry{OMCTS short}

	\addplot[smooth,blue, mark=square*, dashed] plot coordinates { 
		(0.01,0.88000)
		(0.1,0.78000)
		(1,0.84000)
		(10,0.74000)
	};
	\addlegendentry{OMCTS long}

	\addplot[smooth,teal, mark=triangle] plot coordinates { 
		(0.01,0.65000)
		(0.1,0.78000)
		(1,0.76000)
		(10,0.56000)
	};
	\addlegendentry{OMCTS-1 short}

	\addplot[smooth,teal, mark=triangle*, dashed] plot coordinates { 
		(0.01,0.77000)
		(0.1,0.77000)
		(1,0.88000)
		(10,0.80000)
	};
	\addlegendentry{OMCTS-1 long}

	\addplot[smooth,olive, mark=diamond] plot coordinates { 
		(0.01,0.55000)
		(0.1,0.68000)
		(1,0.65000)
		(10,0.53000)
	};
	\addlegendentry{OMCTS-1 short}

	\addplot[smooth,olive, mark=diamond*, dashed] plot coordinates { 
		(0.01,0.76000)
		(0.1,0.64000)
		(1,0.77000)
		(10,0.70000)
	};
	\addlegendentry{OMCTS-1 long}

	\addplot[smooth,red, mark=pentagon] plot coordinates { 
		(0.01,0.63000)
		(0.1,0.69000)
		(1,0.68000)
		(10,0.54000)
	};
	\addlegendentry{OMCTS-3 short}

	\addplot[smooth,red, mark=pentagon*, dashed] plot coordinates { 
		(0.01,0.71000)
		(0.1,0.88000)
		(1,0.74000)
		(10,0.76000)
	};
	\addlegendentry{OMCTS-3 long}

\legend{};
\nextgroupplot[xmode=log,xticklabels={,0,0.1,1,10}, ylabel=score, xlabel=Noise, title = {},]
	\addplot[smooth,black, mark=oplus] plot coordinates { 
		(0.01,7.5400)
		(0.1,6.6100)
		(1,3.8100)
		(10,0.3500)
	};
	\addlegendentry{MCTS short}

	\addplot[smooth,black, mark=oplus*, dashed] plot coordinates { 
		(0.01,6.0800)
		(0.1,6.1300)
		(1,6.2600)
		(10,0.5600)
	};
	\addlegendentry{MCTS long}

	\addplot[smooth,blue, mark=square] plot coordinates { 
		(0.01,8.7900)
		(0.1,8.5200)
		(1,8.4500)
		(10,6.6100)
	};
	\addlegendentry{OMCTS short}

	\addplot[smooth,blue, mark=square*, dashed] plot coordinates { 
		(0.01,9.1300)
		(0.1,8.8300)
		(1,8.6800)
		(10,7.4100)
	};
	\addlegendentry{OMCTS long}

	\addplot[smooth,teal, mark=triangle] plot coordinates { 
		(0.01,8.6300)
		(0.1,8.4500)
		(1,8.5600)
		(10,6.3500)
	};
	\addlegendentry{OMCTS-1 short}

	\addplot[smooth,teal, mark=triangle*, dashed] plot coordinates { 
		(0.01,8.5100)
		(0.1,8.7000)
		(1,8.6900)
		(10,7.0300)
	};
	\addlegendentry{OMCTS-1 long}

	\addplot[smooth,olive, mark=diamond] plot coordinates { 
		(0.01,9.0700)
		(0.1,8.6100)
		(1,8.5900)
		(10,6.3700)
	};
	\addlegendentry{OMCTS-1 short}

	\addplot[smooth,olive, mark=diamond*, dashed] plot coordinates { 
		(0.01,9.3600)
		(0.1,8.8600)
		(1,8.6700)
		(10,7.4500)
	};
	\addlegendentry{OMCTS-1 long}

	\addplot[smooth,red, mark=pentagon] plot coordinates { 
		(0.01,8.9500)
		(0.1,8.7900)
		(1,8.2600)
		(10,6.7500)
	};
	\addlegendentry{OMCTS-3 short}

	\addplot[smooth,red, mark=pentagon*, dashed] plot coordinates { 
		(0.01,9.0500)
		(0.1,9.0100)
		(1,8.5700)
		(10,7.0200)
	};
	\addlegendentry{OMCTS-3 long}

\legend{};
\nextgroupplot[xmode=log,xticklabels={,0,0.1,1,10}, xlabel=Noise, title = {},]
	\addplot[smooth,black, mark=oplus] plot coordinates { 
		(0.01,44.8700)
		(0.1,27.3100)
		(1,9.4500)
		(10,0.8600)
	};
	\addlegendentry{MCTS short}

	\addplot[smooth,black, mark=oplus*, dashed] plot coordinates { 
		(0.01,42.1300)
		(0.1,29.8700)
		(1,19.0300)
		(10,-0.1600)
	};
	\addlegendentry{MCTS long}

	\addplot[smooth,blue, mark=square] plot coordinates { 
		(0.01,59.9000)
		(0.1,55.9100)
		(1,51.5700)
		(10,30.9800)
	};
	\addlegendentry{OMCTS short}

	\addplot[smooth,blue, mark=square*, dashed] plot coordinates { 
		(0.01,67.8200)
		(0.1,64.8500)
		(1,61.1000)
		(10,37.7100)
	};
	\addlegendentry{OMCTS long}

	\addplot[smooth,teal, mark=triangle] plot coordinates { 
		(0.01,46.6800)
		(0.1,54.5700)
		(1,49.4100)
		(10,32.1200)
	};
	\addlegendentry{OMCTS-1 short}

	\addplot[smooth,teal, mark=triangle*, dashed] plot coordinates { 
		(0.01,31.0800)
		(0.1,64.6500)
		(1,61.2200)
		(10,37.5700)
	};
	\addlegendentry{OMCTS-1 long}

	\addplot[smooth,olive, mark=diamond] plot coordinates { 
		(0.01,52.7300)
		(0.1,51.2000)
		(1,48.3800)
		(10,29.7700)
	};
	\addlegendentry{OMCTS-1 short}

	\addplot[smooth,olive, mark=diamond*, dashed] plot coordinates { 
		(0.01,57.8300)
		(0.1,59.1200)
		(1,55.7500)
		(10,33.3500)
	};
	\addlegendentry{OMCTS-1 long}

	\addplot[smooth,red, mark=pentagon] plot coordinates { 
		(0.01,55.2000)
		(0.1,55.5000)
		(1,50.7900)
		(10,32.3200)
	};
	\addlegendentry{OMCTS-3 short}

	\addplot[smooth,red, mark=pentagon*, dashed] plot coordinates { 
		(0.01,50.5000)
		(0.1,65.6500)
		(1,60.9400)
		(10,37.4200)
	};
	\addlegendentry{OMCTS-3 long}

\legend{};
\nextgroupplot[xmode=log,xticklabels={,0,0.1,1,10}, xlabel=Noise, title = {},]
	\addplot[smooth,black, mark=oplus] plot coordinates { 
		(0.01,862.7400)
		(0.1,171.3600)
		(1,4.3900)
		(10,1.6500)
	};
	\addlegendentry{MCTS short}

	\addplot[smooth,black, mark=oplus*, dashed] plot coordinates { 
		(0.01,825.3800)
		(0.1,78.2400)
		(1,90.0400)
		(10,1.9500)
	};
	\addlegendentry{MCTS long}

	\addplot[smooth,blue, mark=square] plot coordinates { 
		(0.01,888.3000)
		(0.1,876.4900)
		(1,824.8500)
		(10,521.8000)
	};
	\addlegendentry{OMCTS short}

	\addplot[smooth,blue, mark=square*, dashed] plot coordinates { 
		(0.01,1006.3200)
		(0.1,952.9800)
		(1,942.0900)
		(10,743.2500)
	};
	\addlegendentry{OMCTS long}

	\addplot[smooth,teal, mark=triangle] plot coordinates { 
		(0.01,826.5000)
		(0.1,910.4000)
		(1,866.3200)
		(10,450.9400)
	};
	\addlegendentry{OMCTS-1 short}

	\addplot[smooth,teal, mark=triangle*, dashed] plot coordinates { 
		(0.01,859.1600)
		(0.1,921.6900)
		(1,985.7400)
		(10,815.6500)
	};
	\addlegendentry{OMCTS-1 long}

	\addplot[smooth,olive, mark=diamond] plot coordinates { 
		(0.01,813.3600)
		(0.1,836.7600)
		(1,802.0600)
		(10,459.5600)
	};
	\addlegendentry{OMCTS-1 short}

	\addplot[smooth,olive, mark=diamond*, dashed] plot coordinates { 
		(0.01,910.0400)
		(0.1,847.2400)
		(1,888.5400)
		(10,740.5700)
	};
	\addlegendentry{OMCTS-1 long}

	\addplot[smooth,red, mark=pentagon] plot coordinates { 
		(0.01,804.9900)
		(0.1,825.3500)
		(1,823.9400)
		(10,460.4900)
	};
	\addlegendentry{OMCTS-3 short}

	\addplot[smooth,red, mark=pentagon*, dashed] plot coordinates { 
		(0.01,848.8200)
		(0.1,973.5600)
		(1,867.7800)
		(10,764.1200)
	};
	\addlegendentry{OMCTS-3 long}

\legend{};
\end{groupplot}
\node at ($(group c2r3) + (0.01,-3.5cm)$) {\ref{grouplegend}}; 
\end{tikzpicture}